\documentclass[11pt]{article}

\usepackage[preprint]{acl}

\usepackage{times}              
\usepackage{latexsym}           
\usepackage[T1]{fontenc}        
\usepackage[utf8]{inputenc}     
\usepackage{microtype}          
\usepackage{inconsolata}        

\usepackage{graphicx}           
\usepackage{subcaption}         

\usepackage{booktabs}           
\usepackage{multirow}           
\usepackage{makecell}           
\usepackage{longtable}          
\usepackage{bigstrut}           
\usepackage[table]{xcolor}      
\usepackage{amsfonts}           
\usepackage{amsmath, mathtools} 
\usepackage{nicefrac}           

\usepackage{enumitem}           

\usepackage{url}                
\usepackage[capitalise,noabbrev]{cleveref} 
\crefname{equation}{Eq.}{Eqs.}

\usepackage[toc,page,header]{appendix}
\usepackage{minitoc}            

\usepackage{caption}            
\usepackage{placeins}           
\usepackage{float}              

\usepackage[most]{tcolorbox}    
\tcbuselibrary{raster}          
\tcbuselibrary{theorems,skins,breakable} 

\usepackage{pifont}             
\usepackage{lipsum}             
\usepackage{multicol}           

\definecolor{darkblue}{HTML}{1A3A5A}      
\definecolor{highlight}{HTML}{F9F7E8}     
\definecolor{emphasis}{HTML}{B30000}      
\definecolor{pale}{HTML}{F5F7FA}          
\definecolor{lightblue}{HTML}{E6EEF7}     
\definecolor{lightgreen}{HTML}{EFF7EE}    
\definecolor{lightgreen2}{RGB}{145, 204, 117} 

\newcommand{\ans}[1]{{\color{emphasis}\textbf{#1}}} 

\newtcolorbox{simplequestion}{
  enhanced,
  breakable,
  colback=pale,
  colframe=darkblue,
  fonttitle=\bfseries\rmfamily,
  coltitle=white,
  boxrule=1pt,
  titlerule=0pt,
  toprule=1pt,
  bottomrule=1pt,
  arc=1mm,
  title={Question},
  halign title=center,
}

\newtcolorbox{answerbox}{
  enhanced,
  breakable,
  colback=pale,
  colframe=darkblue,
  boxrule=0.5pt,
  left=5pt, right=5pt, top=5pt, bottom=5pt,
  arc=1mm,
  notitle,
  width=\textwidth
}

\newtcolorbox{ori-ans}[1][]{
  enhanced,
  breakable,
  colback=lightblue,
  colframe=darkblue,
  fonttitle=\bfseries\rmfamily,
  coltitle=white,
  boxrule=0.5pt,
  left=5pt, right=5pt, top=5pt, bottom=5pt,
  arc=1mm,
  toptitle=1mm,
  bottomtitle=1mm,
  titlerule=0pt,
  halign title=center,
  width=1\textwidth,
  #1  
}

\newtcolorbox{combinedbox}{ 
  enhanced, 
  breakable, 
  colback=pale, 
  colframe=darkblue, 
  fonttitle=\bfseries\rmfamily, 
  coltitle=white, 
  boxrule=1pt, 
  titlerule=0pt, 
  toprule=1pt, 
  bottomrule=1pt, 
  arc=1mm, 
  width=\linewidth,  
}

\newtcolorbox[auto counter, number within=section]{promptbox}[3][Prompt]{
  colback=black!5!white,
  arc=5pt,
  boxrule=0.5pt,
  fonttitle=\bfseries,
  title={#1 \thetcbcounter},  
  before upper={\small\phantomsection},
  fontupper=\fontfamily{ptm}\selectfont,
  colframe=#2,
  label=#3,
}

\newcommand{\modelcomparison}[4]{%
  \begin{combinedbox}%
    \begin{simplequestion}%
      #1%
    \end{simplequestion}%
    \begin{ori-ans}[title={#4}]%
      #2%
    \end{ori-ans}%
  \end{combinedbox}%
}

\newcommand{\MethodName}{OThink-R1}  

\newcommand{\ie}{\emph{i.e.}, }      
\newcommand{\eg}{\emph{e.g.}, }      
\newcommand{\cf}{\emph{cf. }}        

\newcommand{\vx}{\mathbf{x}}         
\newcommand{\vy}{\mathbf{y}}         

\definecolor{customred}{RGB}{192,0,0}      
\definecolor{customblue}{RGB}{31,78,120}   
\definecolor{lightblue}{RGB}{238,245,252}  

\author{
  \textbf{Shengjia Zhang\textsuperscript{1\ding{44}}}, 
  \textbf{Junjie Wu\textsuperscript{2\ding{44}}}, 
  \textbf{Jiawei Chen\textsuperscript{1\ding{41}}}, 
  \textbf{Changwang Zhang\textsuperscript{2\ding{41}}},\\
  \textbf{Zhe Li\textsuperscript{1}}, 
  \textbf{Xingyu Lou\textsuperscript{2}},  
  \textbf{Wangchunshu Zhou\textsuperscript{2}},  
    \textbf{Sheng Zhou\textsuperscript{1}},
    \textbf{Can Wang\textsuperscript{1}},
    \textbf{Jun Wang\textsuperscript{2\ding{41}}}\\  
  \textsuperscript{1} Zhejiang University, \textsuperscript{2} OPPO Research Institute\\
  \small{shengjia.zhang@zju.edu.cn, wujunjie1@oppo.com, sleepyhunt@zju.edu.cn, changwangzhang@foxmail.com}\\
  \small{zheli03@zju.edu.cn, louxingyu@oppo.com, zhouwangchunshu@oppo.com, zhousheng\_zju@zju.edu.cn}\\ \small{wcan@zju.edu.cn, junwang.lu@gmail.com}
}

\title{OThink-R1: Intrinsic Fast/Slow Thinking Mode Switching for Over-Reasoning Mitigation}

\begin{document}
\maketitle

\let\oldthefootnote\thefootnote
\renewcommand{\thefootnote}{}

\footnotemark
\footnotetext{\textsuperscript{\ding{44}}Equal contribution. \textsuperscript{\ding{41}}Corresponding author.}

\let\thefootnote\oldthefootnote

\begin{abstract}
Human cognition operates through two complementary modes: fast intuitive thinking and slow deliberate thinking.  Vanilla large language models (LLMs) predominantly follow the fast‑thinking paradigm, producing immediate responses; while recent large reasoning models (LRMs) adopt slow‑thinking strategies, generating detailed reasoning chains before arriving at answers.  While LRMs often achieve higher accuracy, this comes at the cost of substantially increased token usage. To address this efficiency–accuracy trade‑off, we propose OThink‑R1, a hybrid reasoning framework that integrates both modes within a single LRM and enables automatic mode switching based on problem characteristics. We first identifies three major patterns of essential and redundant reasoning trajectories in LRMs, which guide the design of an auxiliary LLM‑based judge that adaptively determines when slow thinking is necessary. Leveraging the judge’s decisions, we construct a a hybrid fine‑tuning dataset by pruning redundant reasoning to produce fast‑thinking samples and retaining complete reasoning for slow‑thinking samples.  This dataset is then used to fine‑tune LRMs, equipping them with inherent autonomous mode‑selection capabilities. Extensive experiments on mathematical and question‑answering benchmarks show that OThink‑R1 reduces reasoning token usage significantly while maintaining competitive accuracy. The code is available at \url{https://github.com/AgenticIR-Lab/OThink-R1}.
\end{abstract}


\section{Introduction}

Human cognitive processes are often divided into two complementary modes~\citep{kahneman2011thinking,li2025system,WASON1974141,evans2008dual,stanovich2000advancing}: \textbf{1) Fast intuitive thinking} (System 1), which is is instinctive and effortless, allowing for quick decisions but often leading to cognitive biases in complex situations; \textbf{2) Slow deliberate thinking} (System 2), which is thoughtful and analytical, employing logical reasoning to achieve more accurate problem-solving. Traditional large language models (LLMs) primarily mimic fast thinking, relying on heuristics from vast data patterns, which can limit their effectiveness in handling intricate tasks. Recent advancements in language reasoning models (LRMs), such as OpenAI o1~\citep{jaech2024openai} and DeepSeek-R1~\citep{guo2025deepseek}, emulate slow thinking by generating explicit reasoning chains (CoT)~\citep{wei2022chain}, transforming raw predictions into more structured, step-by-step solutions. This approach helps to mitigate the biases of fast thinking, significantly enhancing their reasoning capabilities for tackling complex problems.

\begin{figure*}[t]
    \centering
        \centering
        \includegraphics[width=\textwidth]{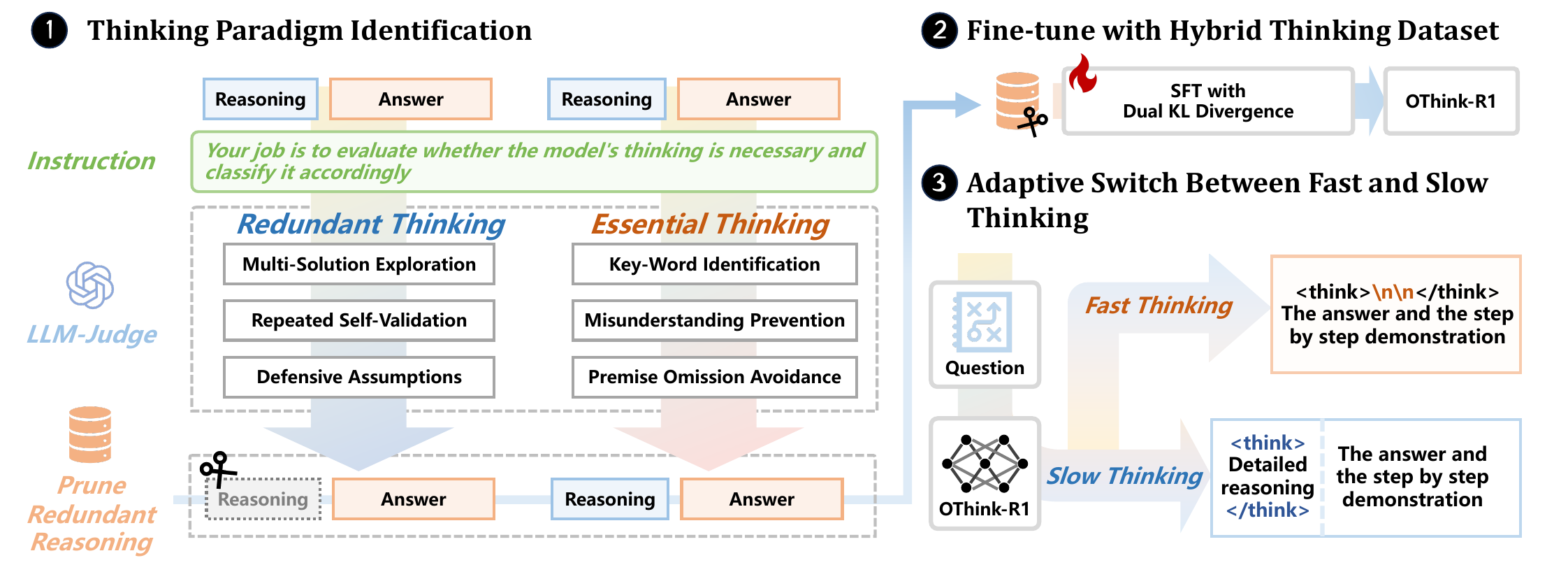}
        \caption{Illustration of the proposed OThink-R1 that equips LRMs with the adaptive hybrid reasoning ability. The pipeline consists of two main steps: \ding{182} \textbf{Thinking Paradigm Identification}. Distinctive patterns differentiating essential from redundant reasoning are extracted from LRM trajectories and organized as prompts to guide an special LLM to act as a judge in classifying reasoning trajectories.   \ding{183} \textbf{Fine-tune with Hybrid Thinking Dataset}. The hybrid dataset is constructed by removing redundant reasoning trajectories to form fast-thinking samples and preserving essential ones as slow-thinking samples. The model is then fine-tuned on this dataset with dual KL-divergence constraint.}
    \label{fig:combined-fig}
\end{figure*}

Despite the impressive performance on complex tasks, LRMs have been associated with considerably higher computational overhead than LLMs. For instance, across commonly solved questions by both model types, typical LRMs generate on average 7.32 times more tokens than non-reasoning LLMs (\cref{tab:overlap}). While several studies have explored token compression techniques to control the length of reasoning for LRMs~\cite{su2024dualformer,hou2025thinkprune}\footnote{While DualFormer~\cite{su2024dualformer} claims to support adaptively switching between fast and slow thinking, this capability was demonstrated exclusively on maze navigation and Sokoban game tasks rather than LLM reasoning tasks.}, these approaches still uniformly employ a slow‑thinking paradigm across all tasks, irrespective of variations in problem difficulty. As a result, relatively straightforward tasks still require excessive reasoning, while complex tasks are constrained by token limits that undermine the completeness of reasoning. This inflexibility introduces an inherent trade‑off between computational efficiency and reasoning completeness, where reductions in token length are frequently accompanied by a decline in accuracy~\cite{hou2025thinkprune}.

To address this issue, we advocate for a new hybrid paradigm that integrates both fast-thinking and slow-thinking modes within a single model, enabling autonomous selection of the appropriate mode for different problems. This concept is inspired by cognitive theories \citep{kahneman2011thinking}, which postulate that humans possess both fast and slow thinking capacities and can flexibly switch between them as needed. Typically, fast thinking is used for a multitude of simple tasks, providing responses swiftly; while slow thinking is reserved for complex problems that demand detailed logical reasoning. We seek to transfer this paradigm to LRMs, allowing them to dynamically adapt their reasoning strategies. The rationality behind this approach is supported by our empirical results (\cref{tab:overlap}), which indicate that slow thinking is not always necessary—vanilla non-reasoning LLMs can solve a substantial portion of tasks with significantly fewer tokens. However, implementing adaptive hybrid reasoning in LRMs presents two key challenges: \textit{1) How to identify which scenarios that fast thinking is sufficient?  2) How to empower LRMs to autonomously select the appropriate reasoning mode?} While a few pioneering works, such as Dualformer~\citep{su2024dualformer} and Qwen3~\cite{yang2025qwen3technicalreport}, have explored hybrid modes, their methods still require manual selection of the reasoning mode and lack automatic adaptation.

In this work, we propose \MethodName, a novel framework that enables LRMs to automatically switch between fast and slow thinking modes. To identify instances where fast thinking is sufficient, we systematically analyze LRM  reasoning trajectories and extract distinctive patterns that differentiate essential from redundant reasoning. These patterns are then organized as prompts that guide a auxiliary LLM (\eg GPT‑4o) to act as a judge, classifying reasoning trajectories as either essential or redundant. This classification serves as the basis for training LRMs to perform adaptive reasoning.

To further empower LRMs with adaptive hybrid reasoning capabilities, we explicitly construct a supervised fine‑tuning (SFT) dataset. For tasks in which fast‑thinking LLMs produce correct answers and the LRM’s reasoning trajectories are judged redundant, we remove the redundant steps and retain only the immediate responses, forming the fast‑thinking samples. Conversely, for tasks judged as essential, or where only LRMs produce correct answers, the full reasoning processes are preserved as slow‑thinking samples. Fine‑tuning LRMs on this hybrid dataset enables the model to adaptively select fast or slow thinking modes based on the nature of the problem. Furthermore, we introduce a dual KL‑divergence constraint to prevent thinking mode collapse by anchoring the model to both the original LRM and a non‑reasoning LLM. Extensive experiments across multiple benchmarks demonstrate that \MethodName\ reduces computational overhead while preserving competitive accuracy. Our main contributions are summarized as follows:

\begin{itemize}[leftmargin=*]
\item We propose \MethodName, a hybrid reasoning framework that enables LRMs adaptively switch between fast and slow thinking modes, which is more aligned with human cognitive process.

\item We uncover three major patterns of essential and redundant reasoning trajectories in LRMs, which inform the design of an LLM-based judge for adaptively assessing reasoning necessity.

\item We conduct extensive experiments to validate the effectiveness of the proposed \MethodName, which significantly reduces token cost while maintaining accuracy.

\end{itemize}

\begin{table*}[htbp]
    \scriptsize 
    \centering

    \begin{tabular}{l ccl ccl}
    \Xhline{1.2pt}
    \multirow{3}{*}{\textbf{Dataset}} 
    & \multicolumn{3}{c}{\textbf{7B}} 
    & \multicolumn{3}{c}{\textbf{14B}} 
    \bigstrut[t]\\
    \cmidrule(lr){2-4} \cmidrule(lr){5-7}
    & \textbf{ Qwen Tokens } & \textbf{ R1 Tokens } & \textbf{Overlap Ratio}
    & \textbf{ Qwen Tokens } & \textbf{ R1 Tokens } & \textbf{Overlap Ratio}
    \\
    \midrule

    \multirow{2}{*}{OpenBookQA} 
    & \multirow{2}{*}{36}   & \multirow{2}{*}{638}   & \multirow{2}{*}{\textbf{60.92\%}  (3020/4957)} 
    & \multirow{2}{*}{89}   & \multirow{2}{*}{559}   & \multirow{2}{*}{\textbf{73.01\%} (3619/4957)} 
    \\ 
    & &  & & &  &   \\ 

    \multirow{2}{*}{CommonsenseQA} 
    & \multirow{2}{*}{50}   & \multirow{2}{*}{741}    & \multirow{2}{*}{\textbf{51.20\%} (3990/7793)}  
    & \multirow{2}{*}{65}   & \multirow{2}{*}{572}    & \multirow{2}{*}{\textbf{61.62\%} (4802/7793) } \\ 
    & &  &   & &  &  \\

    \multirow{2}{*}{ASDIV} 
    & \multirow{2}{*}{104} & \multirow{2}{*}{311} & \multirow{2}{*}{\textbf{90.10\%} (865/960)}  
    & \multirow{2}{*}{104} & \multirow{2}{*}{316} & \multirow{2}{*}{\textbf{88.75\%}  (852/960)} \\ 
    & & &   & & &   \\ 

    \multirow{2}{*}{GSM8K} 
    & \multirow{2}{*}{276}  & \multirow{2}{*}{691} &  \multirow{2}{*}{\textbf{87.36\%}  (5223/5979)}
    & \multirow{2}{*}{274}  & \multirow{2}{*}{649} &  \multirow{2}{*}{\textbf{89.31\%}  (5340/5979)} \\  
    & &  &  & &  & \bigstrut[b]\\

    \Xhline{1.2pt}
    \end{tabular}
    \caption{Comparison experiments between typical non-reasoning LLMs (Qwen2.5-Instruct) and LRMs (DeepSeek-R1-Distill-Qwen). In this table, we report: 1) the average number of tokens generated by LLMs and LRMs; 2) the overlap ratio, defined as the proportion of problems correctly solved by both models with respect to all problems.}
    \label{tab:overlap}
\end{table*}


\section{Preliminaries}
In this section, we present the two reasoning modes in human cognition and introduce LLMs and LRMs that correspond to these modes.

\noindent\textbf{Large Language Models (LLMs) and Fast Thinking.} 
Human fast thinking~\citep{kahneman2011thinking,evans2008dual} is characterized by rapid, intuitive responses without deliberate analysis. Traditional LLMs like Qwen-2.5~\citep{qwen2025qwen25technicalreport}, exhibit analogous fast-thinking behavior by directly generating answers without producing intermediate reasoning steps. Formally, given an input $\mathbf{x}$, these models produce output tokens $\mathbf{y}=[y_1,\ldots,y_m]$ based on the token-wise generative probability:
\begin{equation}\label{eqs:generation-paradigm} 
\pi_{\theta}(\mathbf{y}|\mathbf{x})=\prod_{j=1}^{m}\pi_{\theta}(y_j|\mathbf{x},y_{<j}). 
\end{equation} 
This paradigm would direct generate the response as the fast-thinking procedure. However, the lack of explicit reasoning steps leads to suboptimal performance on complex problems~\cite{guo2025deepseek}.

\noindent\textbf{Large Reasoning Models (LRMs) and Slow Thinking.} 
Human slow thinking~\citep{kahneman2011thinking,evans2008dual} engages deliberate analysis through step-by-step reasoning and careful reflection on intermediate results. LRMs replicate this sophisticated reasoning mode by generating explicit reasoning trajectories before drawing final answers. Representative models include OpenAI o1~\citep{jaech2024openai} and DeepSeek-R1~\citep{guo2025deepseek}. These models delineate reasoning processes using special tokens (\eg \texttt{<think>} and \texttt{</think>}).  Within these delimiters, the generated sequences systematically decompose problems into sub-components, explore alternative solution strategies, perform necessary computations, and validate intermediate results. This explicit and sophisticated reasoning paradigm enhances performance on complex tasks, such as mathematical problem solving and question answering. Nonetheless, the production of such reasoning trajectories substantially increases inference latency and computational overhead~\cite{yang2025qwen3technicalreport}.

\noindent\textbf{Toward Adaptive Hybrid Reasoning.} Both vanillia LLMs and LRMs are typically restricted to a single reasoning paradigm, each with its respective limitations:
(1) In LLMs, reliance solely on fast thinking is insufficient for effectively addressing complex problems.
(2) In LRMs, the uniform application of slow thinking to simple tasks leads to unnecessary computational overhead. This inflexibility contrasts with human cognition, where individuals autonomously adjust their reasoning mode according to task difficulty~\citep{kahneman2011thinking,evans2008dual}. Therefore, a more promising approach is to integrate both slow and fast thinking and enable adaptive switching between them according to the problem characteristics.

While recently has seen several concurrent explorations of hybrid reasoning, these approaches exhibit notable limitations. Qwen3~\citep{yang2025qwen3technicalreport} requires users to manually pre‑specify the reasoning mode, and therefore fail to achieve adaptive mode selection.  In contrast, GPT‑5 and Claude-3.7-Sonnet~\citep{Sonnet3_7} claim to enable adaptive switching, but they are closed‑source, whose underlying mechanisms and procedural workflows remain poorly understood.

\section{Methodology}\label{sec:methodology}

In this section, we propose OThink‑R1 that equips LRMs with adaptive hybrid reasoning capabilities. OThink‑R1 comprises two main steps:

\subsection{Thinking Paradigm Identification}
While slow‑thinking LRMs exhibit strong reasoning capabilities, our analysis in \Cref{tab:overlap} shows that both fast‑thinking LLMs (\eg Qwen2.5 series~\citep{qwen2025qwen25technicalreport}) and slow‑thinking LRMs produce correct solutions for a substantial portion of problems—over 50\% across four representative datasets. Notably, fast‑thinking LLMs achieve these results using significantly fewer tokens. This substantial overlap indicates that in many cases, the reasoning processes generated by LRMs do not facilitate concluding correct solutions. This motivates us to identify patterns that characterize reasoning facilitating problem-solving.

\begin{table*}[t]
    \centering
    \resizebox{\textwidth}{!}{
    \begin{tabular}{l cc cc cc cc}
    \Xhline{1.3pt}
    \multirow{3}{*}{\textbf{Model}} & 
    \multicolumn{2}{c}{\textbf{OpenBookQA}} & 
    \multicolumn{2}{c}{\textbf{CommonsenseQA}} & 
    \multicolumn{2}{c}{\textbf{ASDIV}} & 
    \multicolumn{2}{c}{\textbf{GSM8K}} \bigstrut[t]\\
    \cmidrule(lr){2-3} \cmidrule(lr){4-5} \cmidrule(lr){6-7} \cmidrule(lr){8-9}
    & \textbf{Tokens $\downarrow$} & \textbf{ACC $\uparrow$} & 
    \textbf{Tokens $\downarrow$} & \textbf{ACC $\uparrow$} & 
    \textbf{Tokens $\downarrow$} & \textbf{ACC $\uparrow$} & 
    \textbf{Tokens $\downarrow$} & \textbf{ACC $\uparrow$}   \bigstrut\\
    \Xhline{1.0pt}
    DeepSeek-R1-Distill-Qwen-7B  &783 	&76.40\%	 &730 	&64.70\%	 &352 	&97.00\%	 &719 	&86.10\%    \bigstrut[t]\\
    NoThinking-R1-7B~\citep{ma2025reasoning}                      & 130 	& 56.60\%	 &106 	&54.30\%	 &138 	&88.00\%	 &258 	&77.50\%    \bigstrut\\
    DualFormer-R1-7B~\citep{su2024dualformer}                      & 723 	& 75.20\%	 &701 	&66.70\%	 &223 	&96.00\%	 &460 	&81.30\%    \bigstrut\\
    \MethodName-7B                                   &{667} 	&{\textbf{76.80\%}}	 &{634} 	&{\textbf{66.90\%}}	 & {270}  	& {\textbf{98.00\%}} 	 &{488} 	& {\textbf{86.70\%}}   \\
        \midrule
    DeepSeek-R1-Distill-Qwen-14B                      &522 	&92.80\%	 &569 	&81.70\%	 &319 	&97.00\%	 &657 	&91.20\%    \bigstrut[t]\\
    NoThinking-R1-14B~\citep{ma2025reasoning}                      & 296 	& 87.80\%	 &373 	&78.90\%	 &197 	&94.40\%	 &458 	&88.60\%    \bigstrut\\
    DualFormer-R1-14B~\citep{su2024dualformer}                      & 1688 	& 91.40\%	 &2003 	&79.80\%	 &223 	&95.70\%	 &482 	&86.40\%    \bigstrut\\
    \MethodName-14B                                   &{421} 	&{\textbf{93.40\%}}	         &{435} 	&{\textbf{81.80\%}}		         & {412}  	&{\textbf{98.00\%}}	         &{791}  	& {89.80\%}          \\
    \Xhline{1.3pt}
    \end{tabular}
        }
    \caption{Performance comparison across multiple datasets ($\downarrow$: fewer tokens, better efficiency; $\uparrow$: higher accuracy, better performance). The bolded results indicate that \MethodName\ achieve the state-of-the-art performance.}
    \label{tab:main_results}
\end{table*}

\begin{table*}[htbp] 
    \centering
    \begin{minipage}{0.45\linewidth}
        \scriptsize
        \caption*{\textbf{(1) Prune Ratio (Training).}} 
        \centering
        \begin{tabular}{l cc}  
            \Xhline{1.2pt} 
            \multirow{2}{*}{\textbf{Dataset}} & \multicolumn{2}{c}{\textbf{Prune Ratio (Training)}}\bigstrut[t]\\ 
            \cmidrule(lr){2-3} 
            & \textbf{7B} & \textbf{14B} \\ 
            \midrule 
            \multirow{2}{*}{OpenBookQA} & 75.18 \% & 67.36 \% \\ 
            & (2845/3784) & (3021/4485) \\ 
            \multirow{2}{*}{CommonsenseQA} & 74.89\% & 62.70\%\\ 
            & (3684/4919) & (3870/6172) \\ 
            \multirow{2}{*}{ASDIV} & 32.51\% & 31.45\% \\ 
            & (303/932) & (295/938) \\ 
            \multirow{2}{*}{GSM8K} & 32.11\% & 31.65\%\\ 
            & (1734/5400) & (1738/5492) \\ 
            \Xhline{1.2pt} 
        \end{tabular} 
    \end{minipage} 
    \hspace{0.05\linewidth}
        \begin{minipage}{0.45\linewidth} 
        \scriptsize 
        \centering
        \caption*{  \textbf{(2) Fast-thinking Ratio (Test).}}
        \begin{tabular}{l cc }
            \Xhline{1.2pt}
            \multirow{3}{*}{\textbf{Dataset}} &  \multicolumn{2}{c}{\textbf{Fast-thinking Ratio (Test)}}\bigstrut[t]\\
            \cmidrule(lr){2-3}  &  \textbf{7B} &  \textbf{14B} \\
            \midrule 
            \multirow{2}{*}{OpenBookQA} & \multirow{2}{*}{6.40\%} & \multirow{2}{*}{36.80\%}   \\   & & \\ 
            \multirow{2}{*}{CommonsenseQA} & \multirow{2}{*}{8.76\%} & \multirow{2}{*}{35.70\%}\\   & & \\ 
            \multirow{2}{*}{ASDIV}  & \multirow{2}{*}{13.62\%} & \multirow{2}{*}{7.97\%}\\ & &\\ 
            \multirow{2}{*}{GSM8K}  & \multirow{2}{*}{9.93\%} & \multirow{2}{*}{8.56\%}\\ & &\\ 
            \Xhline{1.2pt}
        \end{tabular}
    \end{minipage}%
    \caption{Prune and fast-thinking ratios across model scales and datasets. In this table, we report: (1) \textbf{Prune Ratio (Training)}: the proportion of training examples where redundant reasoning trajectories within \texttt{<think>} were pruned; (2) \textbf{Fast-thinking Ratio (Test)}: the proportion of test examples where the model activates fast-thinking.}
    \label{tab:fast-thinking-ratio} 
\end{table*}

To address this, we examine problems where LRMs succeed but LLMs fail: in these cases, reasoning demonstrably enables correct solutions that are unattainable without it. To extract characteristics of such facilitative reasoning, we invited over 10 senior NLP researchers to analyze 100 reasoning trajectories from these problems. Researchers identified common patterns across these trajectories and provided explanations for why these patterns characterize essential reasoning (examples are provided in \Cref{appendix:essential_reasoning}). Synthesizing these expert analyses, we identify trajectories exhibiting these patterns as \textbf{essential reasoning}, characterized by the following properties:

\begin{enumerate}[leftmargin=*] 
    \item \textbf{Key-word Identification:} 
    Extracting and emphasizing critical problem components forms the basis for solution derivation. 

    \item \textbf{Misunderstanding Prevention:} 
    Eliminating misunderstandings in the problem statement prevents errors from incorrect assumptions. 
    
    \item \textbf{Premise Omission Avoidance:} 
    Comprehensive coverage of all given premises and conditions ensures solution validity.
\end{enumerate}

Conversely, we also invited the same senior NLP researchers to examine 100 reasoning trajectories from problems where LLMs succeed but LRMs fail—in these cases, reasoning leads to incorrect solutions while direct generation succeeds. Common patterns identified across these trajectories characterize \textbf{redundant reasoning}:
\begin{enumerate}[leftmargin=*] 
    \item \textbf{Multi-Solution Exploration:} Persistent exploration of alternative solutions despite determining a correct answer.   
    
    \item \textbf{Repeated Self-Validation:} Excessive re-validation of every intermediate step in final solutions. 
         
    \item \textbf{Defensive Assumptions:} Being overly cautious, taking extraneous hypotheses into consideration based on internal knowledge, rather than problem-specific constraints. 
\end{enumerate}

\subsection{Fine-tune with Hybrid Thinking Dataset}

Having identified characteristics of essential and redundant reasoning, the next question is how to equip LRMs with the ability to autonomously select the appropriate reasoning mode. A straightforward idea is to construct a supervised fine-tuning (SFT)  dataset that teaches LRMs to  choose the proper reasoning mode (fast or slow) for a given problem.

\noindent\textbf{Constructing Hybrid SFT Dataset.} To construct a dataset that enables adaptive reasoning, the identified essential and redundant patterns are organized as prompts that guide an LLM (GPT-4o) to act as a judge, classifying LRM reasoning trajectories as essential or redundant (\cf \Cref{appendix:system_prompt}). Based on this classification, we construct a dataset consisting of two components:

\begin{itemize}[leftmargin=*]
    \item \textbf{Fast-Thinking Sub-Dataset}: We curate this dataset from problems where both LRMs and LLMs produce correct solutions and reasoning is classified as redundant. For these problems, the reasoning content is removed while the immediate conclusions are retained, creating fast-thinking samples.
    
    \item \textbf{Slow-Thinking Sub-Dataset}: We curate this dataset from two sources: (1) problems where both models succeed but reasoning is classified as essential, and (2) problems where only LRMs succeed. In both cases, complete reasoning trajectories are preserved.
\end{itemize}

The constructed dataset $\mathcal{D}_{\text{train}}$ contains both fast-thinking (immediate responses) and slow-thinking (detailed reasoning) samples. By exposing the LRM to both types during fine-tuning, it learns to adaptively engage in step-by-step reasoning, or directly drawing conclusions based on problem characteristics.

\noindent\textbf{Fine-tune with Dual KL-Divergence Loss.} Direct fine-tuning on $\mathcal{D}_{\text{train}}$ using standard maximum likelihood estimation poses a challenge: the dataset contains two fundamentally different generation patterns. Fast-thinking samples resemble non-reasoning LLMs' direct response style, while slow-thinking samples follow LRMs' deliberate reasoning style. This diversity risks the model collapsing to a single pattern during training, as the model tends to prioritize one generation pattern over the other. Empirical results (\Cref{sec:ablation-study}) show that standard training leads to performance degradation.

To address this issue, we incorporate dual KL-divergence constraints that anchor the fine-tuned model to both the reference LRM and LLM:
\begin{equation}\label{eqs:hybrid-loss}
\begin{aligned}
    &\mathcal{L} =\mathbb{E}_{(\vx,\vy)\in\mathcal{D}_{\text{train}}}\Big[\beta_1 \text{D}_{\text{KL}}(\pi_{\theta}\Vert\pi_{\text{LRM}};\vx,\vy) \\
    &+ \beta_2  \text{D}_{\text{KL}}(\pi_{\theta}\Vert\pi_{\text{LLM}};\vx,\vy)  
    -\log\pi_{\theta}(\vy\vert \vx)
    \Big]
\end{aligned}
\end{equation}
where $\pi_{\text{LRM}}$ and $\pi_{\text{LLM}}$ denote the reference LRM and LLM respectively. $\pi_{\theta}(\vy\vert\vx)$ is defined in \Cref{eqs:generation-paradigm}. The regularization term $\beta_1$ preserves reasoning capability while $\beta_2$ maintains efficient generation, enabling the model to balance between fast and slow thinking. Let $\mathcal{V}$ denotes the token vocabulary, the KL term $\text{D}_{\text{KL}}(\pi_{1}\Vert\pi_{2};\vx,\vy) $ is:
\begin{equation}
    \frac{1}{\vert\vy\vert}\sum_{j=1}^{\vert \vy \vert} \sum\limits_{v\in\mathcal{V}}\pi_1(v\vert\vx,y_{<j})\log\frac{\pi_1(v\vert\vx,y_{<j})}{\pi_2(v\vert\vx,y_{<j})}
\end{equation}

\begin{table*}[htbp]
    \tiny
    \centering

    \label{tab:ablation_study}
    \resizebox{\textwidth}{!}{
    \begin{tabular}{l cc cc cc cc}
    \Xhline{1.2pt}
    \multirow{3}{*}{\textbf{Model}} & 
    \multicolumn{2}{c}{\textbf{OpenBookQA}} & 
    \multicolumn{2}{c}{\textbf{CommonsenseQA}} & 
    \multicolumn{2}{c}{\textbf{ASDIV}} & 
    \multicolumn{2}{c}{\textbf{GSM8K}} \bigstrut[t]\\
    \cmidrule(lr){2-3} \cmidrule(lr){4-5} \cmidrule(lr){6-7} \cmidrule(lr){8-9}
    & \textbf{Tokens $\downarrow$} & \textbf{ACC $\uparrow$} & 
    \textbf{Tokens $\downarrow$} & \textbf{ACC $\uparrow$} & 
    \textbf{Tokens $\downarrow$} & \textbf{ACC $\uparrow$} & 
    \textbf{Tokens $\downarrow$} & \textbf{ACC $\uparrow$}   \\
    \midrule
    w/o prune (7B)                      &997  	&75.40\%	 & 775  	&66.10\%	&249 	&98.00\% &432 	&86.40\%   \\
   w/o LLM-Judge (7B)                   &997&69.40\%&776&65.70\%&171&97.00\%&279&84.20\% \\
    w/o KL-constraint (7B)              &328  	&68.80\%	 & 268  	&64.00\%	 &292 	&97.30\% &588 	&86.40\%   \\
    w/o reference LRM (7B)              & 4364&72.20\%&4320&69.40\%&2144&97.70\%&2073&84.20\% \\
    w/o reference LLM (7B)              & 649&76.40\%&617&64.20\%&267&97.30\%&480&85.50\% \\
    \MethodName-7B                                                       & {667  }&{ 76.80\%}	 &{ 634  	}&{ 66.90\% }	    & {270 } 	                & {98.00\%} 	 &{ 488 } 	        & {86.70\%}    \bigstrut[b]\\
    \Xhline{1.0pt}
    w/o prune  (14B)                     &1676  	&94.60\%	 & 1455  	&80.70\%	     &305 	&96.70\% &629 	&91.70\%  \bigstrut[t]\\
   w/o LLM-Judge (14B)                   &4353&93.20\%&5038&81.50\%&196&98.00\%&300&88.70\% \\
    w/o KL-constraint (14B)              &3731  	&94.80\%	 & 3001  	&81.60\%	     &281 	&97.00\% &338 	&91.10\%  \\
    w/o reference LRM (14B)              & 10599&89.00\%&11516&80.80\%&8566&89.70\%&4807&82.70\% \\
    w/o reference LLM  (14B)             &  253&91.40\%&409&81.20\%&282&96.30\%&475&88.30\% \\
    \MethodName-14B                      &{421} 	&{ 93.40\% }	         &{435} 	&{81.80\%}	         & {412}  	&{98.00\%}	         &{791}  	& {89.80\%}           \bigstrut[b]\\
    \Xhline{1.3pt}
    \end{tabular}
        }
    \caption{Ablation study across multiple datasets. ($\downarrow$: lower is better, $\uparrow$: higher is better)}
\end{table*}

\section{Experiments}

\subsection{Experimental Setting}\label{sec:implementation-details}
 
 \textbf{Datasets.} We conduct experiments on four widely-used datasets: \textbf{OpenBookQA}~\cite{OpenBookQA2018} and \textbf{CommonsenseQA}~\cite{talmor-etal-2019-commonsenseqa}, two commonly used datasets for question answering evaluation; \textbf{ASDIV}~\cite{miao2021diverse} and \textbf{GSM8K}~\cite{cobbe2021trainingverifierssolvemath}, two commonly used datasets for mathematical reasoning evaluation. We refer readers to \cref{appendix:experiment_datasets} for more details.

\noindent\textbf{Models.} We fine-tune the DeepSeek-R1-Distill-Qwen series (\ie 7B/14B variants) on the constructed hybrid thinking dataset.  To align with the fine-tuned model, we use the same DeepSeek-R1-Distill-Qwen series as the reference LRM ($\pi_{\text{LRM}}$ in \Cref{eqs:hybrid-loss}); Additionally, we select Qwen2.5-Instruct series, the non-reasoning model that directly generates immediate responses, as the reference LLM ($\pi_{\text{LLM}}$ in \Cref{eqs:hybrid-loss}). Both reference models match the size of the fine-tuned model.

\noindent\textbf{Baselines.} We compare two baselines: 1) \textbf{NoThinking} \citep{ma2025reasoning}, a training-free method which directly bypass the reasoning process with explicit prompting; 2) \textbf{DualFormer} \citep{su2024dualformer} proposed dropping reasoning steps during training, which enables thinking mode switch on structured tasks (\eg maze navigation and Sokoban). However, in LLM reasoning tasks, the DualFormer fails to achieve the adaptive hybrid reasoning. In this paper, we follow its LLM setting by randomly dropping intermediate reasoning steps in large reasoning model outputs to construct the fine-tuning dataset, and then conduct SFT.

 \noindent\textbf{Hyperparameter Tuning.}  A grid search is adopted to find optimal hyperparameters. For all compared methods, we closely follow configurations in their respective publications to ensure the optimal performance. For more details, we refer readers to \cref{appendix:optimal_hyperparameters}.

\subsection{Main Results}

 \noindent\textbf{Performance comparison on QA tasks.} As shown in \cref{tab:main_results}, OThink-R1 achieves superior accuracy while generating fewer tokens on OpenBookQA and CommonsenseQA at both parameter scales. NoThinking~\citep{ma2025reasoning} instructs the model to skip reasoning by incorporating the special token (\texttt{</think>}) into the prompt, which generates the fewest tokens but suffers severe performance degradation compared to the base LRM. This premature termination disrupts the model's reasoning structure and leads to substantial accuracy drop. DualFormer~\citep{su2024dualformer} fails to achieve adaptive hybrid reasoning, exhibiting unstable performance across benchmarks. Notably, at the 14B scale, DualFormer substantially increases token counts on QA tasks while decreasing performance relative to the base LRM. In contrast, \MethodName\ successfully achieves adaptive hybrid reasoning (\cf \Cref{tab:fast-thinking-ratio}), and also improves both accuracy and efficiency on QA tasks.

\noindent\textbf{Performance comparison on mathematical reasoning tasks.} On mathematical reasoning tasks, \MethodName\ maintains competitive or superior accuracy compared to the base LRM and baseline methods. At the 7B scale, \MethodName\ achieves the highest accuracy on both tasks while generating fewer tokens compared with the base LRM. In addition, \MethodName\ also demonstrates adaptive hybrid reasoning capability (\cf \cref{tab:fast-thinking-ratio}).

\begin{figure*}[t]
    \centering
        \centering
\includegraphics[width=\textwidth]{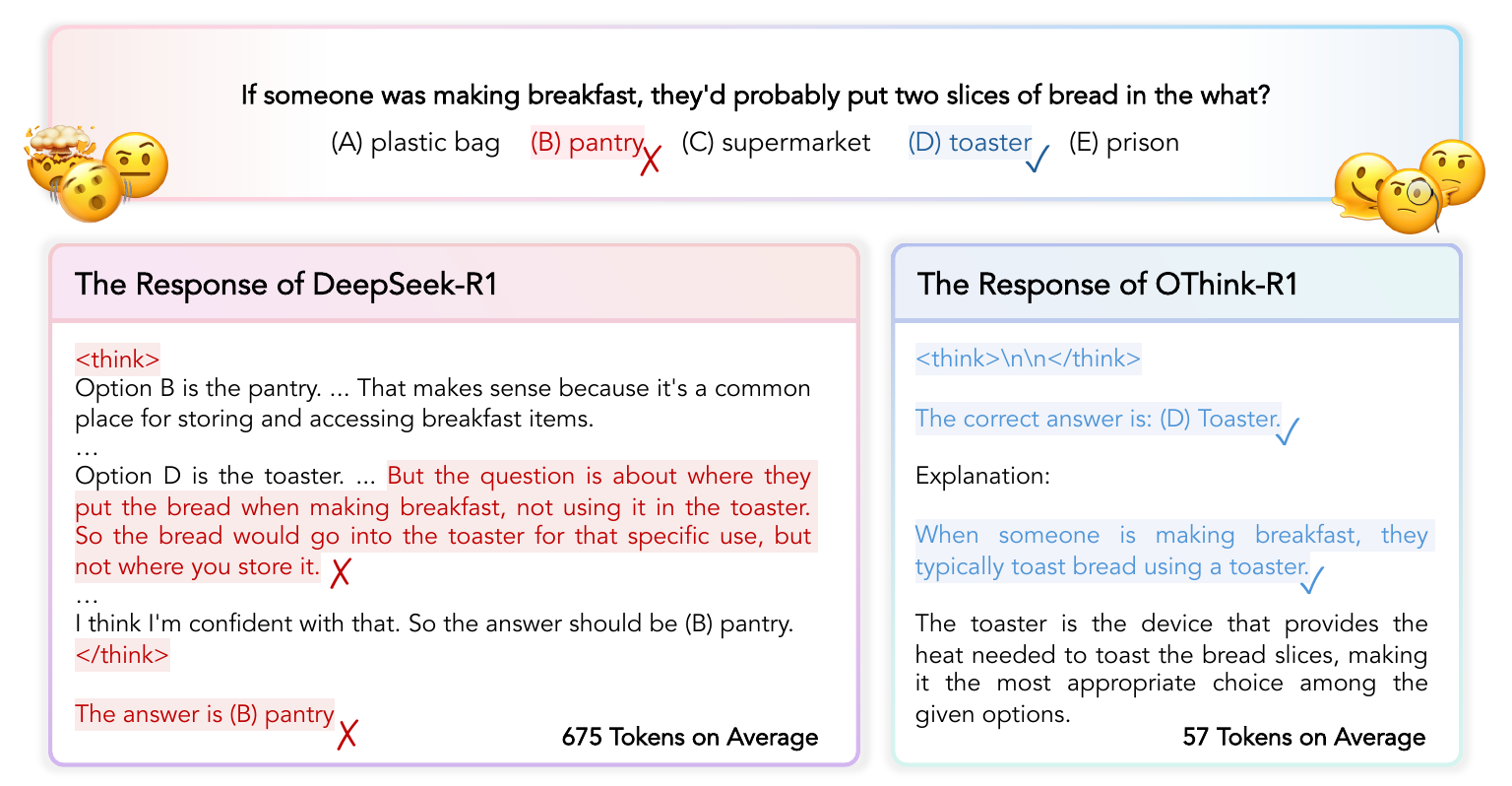}
        \caption{Comparison of DeepSeek-R1-Distill-Qwen-7B/OThink-R1-7B generated responses on CommonsenseQA.}
    \label{fig:case_study-Official}
\end{figure*}

\subsection{Ablation Study}\label{sec:ablation-study}

In this section, five variants of \MethodName\ were evaluated: 1) \textbf{w/o prune}. \MethodName\ selectively removes redundant reasoning trajectories while preserving essential ones. In contrast, this variant retains reasoning trajectories for all training samples without any pruning; 2) \textbf{w/o LLM-Judge}. \MethodName\ employs an auxiliary LLM to classify reasoning as essential or redundant, removing only redundant cases. This variant removes the LLM-Judge component and treats all cases where both LRM and LLM answers are correct as redundant, removing their reasoning trajectories; 3) \textbf{w/o KL-constraint}, which employed vanilla SFT loss without any KL divergence constraint; 4) \textbf{w/o reference LRM} ($\beta_1=0$), which removed the KL constraint on the reference LRM while retaining the LLM constraint; 5) \textbf{w/o reference LLM} ($\beta_2=0$), which removed the KL constraint on the reference LRM while retaining the LRM constraint.

\noindent\textbf{The benefits of hybrid reasoning.} \MethodName\ outperforms the w/o prune variant across most benchmarks. Since w/o prune retains all reasoning trajectories and relies solely on slow thinking, the superior performance validates the effectiveness of the hybrid reasoning.

\noindent\textbf{Reasoning classification matters.} The w/o LLM-Judge variant underperformed \MethodName\ across all datasets. This demonstrates that answer correctness alone is insufficient---even when both models produce correct answers, some reasoning trajectories remain essential and should be preserved through fine-grained classification. This further validates the necessity of distinguishing essential from redundant reasoning.

\noindent\textbf{The necessity of Dual KL constraints.} \MethodName\ generally outperforms the w/o KL-constraint variant, validating the importance of the dual constraint mechanism. Moreover, the w/o reference LRM variant ($\beta_1=0$) leads to severe overthinking, with token counts significantly increasing (\eg 421 to 10,599 in OpenBookQA on 14B model). In addition, the w/o reference LLM variant ($\beta_2=0$) exhibits performance degradation. Thus, both constraints are essential for balancing efficiency and accuracy.

\subsection{Case Study}\label{sec:case_study}
We present a case study in \Cref{fig:case_study-Official}. \MethodName\ employs fast-thinking mode and solves the problem correctly with an immediate response. In contrast, the baseline LRM exhibits overthinking: despite deriving the correct answer during reasoning, it continues the reasoning process and ultimately conclude the incorrect answer. This demonstrates that \MethodName\ can dynamically switch between fast and slow thinking modes. Moreover, this case illustrates that extensive reasoning is not always beneficial and may even degrade model performance.


\section{Related Work}

\noindent\textbf{Large Reasoning Models.} Recent advances in large reasoning models (LRMs) have demonstrated remarkable capabilities in complex problem-solving tasks. LRMs such as OpenAI o1~\citep{jaech2024openai} and DeepSeek-R1~\citep{guo2025deepseek} generate extended reasoning trajectories that involve step-by-step analysis and reflection on intermediate results before drawing final conclusions. Unlike traditional large language models (LLMs) such as Qwen-2.5 series~\citep{qwen2025qwen25technicalreport} and GPT-4~\citep{achiam2023gpt}, which directly produce answers without explicit reasoning steps, LRMs actively construct longer chains of thought to tackle challenging problems that require multi-step inference and logical deduction.

However, existing LRMs lack the capability to adaptively select appropriate thinking modes. LRMs like OpenAI o1~\citep{jaech2024openai} and DeepSeek-R1~\citep{guo2025deepseek} uniformly apply slow-thinking mode across all tasks, generating reasoning trajectories even for simple questions where direct answers suffice. This uniform application introduces unnecessary computational overhead and latency. While some LRMs attempt to address this limitation, they remain inadequate. Qwen3~\citep{yang2025qwen3technicalreport} requires users to manually specify the reasoning mode before inference, failing to achieve adaptive mode selection. Conversely, closed-source LRMs such as GPT-5 and Claude-3.7-Sonnet~\citep{Sonnet3_7} claim to support adaptive mode switching, but their mechanisms remain unclear and inaccessible for systematic investigation.

\noindent\textbf{Efficient Reasoning.} Recent work reduces computational cost by uniformly shortening reasoning trajectories across all inputs. A common approach applies reinforcement learning with length penalties that discourage longer reasoning chains~\citep{arora2025training,yi2025shorterbetterguidingreasoningmodels,hou2025thinkprune}. To enable more flexible control, L1~\citep{aggarwal2025l1} trains models to match reasoning length with user pre-specified reasoning length, enabling length adjustment for efficient reasoning based on user needs. Alternatively, DAST~\citep{shen2025dast} employs direct preference optimization on length-preference pairs, where shorter correct responses are preferred over longer ones. Other approaches learn compressed representations~\citep{hao2024training,cheng2024compressed,shen2025efficient,xu2025softcot,saunshi2025reasoning}, distill answers from reasoning models to non-reasoning models~\citep{kang2025c3ot}, merge reasoning and non-reasoning models to control output length~\citep{ma2025cot}, use prompt engineering~\citep{han2024token,xu2025chain,renze2024benefits,wu2025effectively}, or prune reasoning steps by semantic importance~\citep{xia2025tokenskip}. Beyond these length-control approaches, some methods dynamically terminate reasoning by detecting failure signals during generation~\citep{yang2025dynamic}.

However, these methods apply uniform strategies across all problems and cannot adaptively switch between fast and slow-thinking modes. his limitation creates a fundamental trade-off: excessive reduction in reasoning length degrades accuracy on complex tasks~\citep{arora2025training,hou2025thinkprune}, while insufficient reduction still incurs unnecessary reasoning costs on simple tasks that could be solved directly.  In contrast, \MethodName enables adaptive hybrid reasoning by identifying when reasoning is necessary, preserving full reasoning capacity on complex problems while bypassing reasoning entirely on simple ones.


\noindent\textbf{Hybrid Reasoning.} Recent works have explored various strategies to equip language models with hybrid reasoning capabilities.  Several studies have proposed routing methods that rely on a base language model~\citep{chuang2025confident,mahaut2024factual} to select between fast and slow-thinking models, either by training it as a router~\citep{ong2024routellm} or by analyzing its output distributions~\citep{su2025cp}. However, this reliance on an external language model incurs substantial parameter overhead. In contrast, \MethodName\ equips the reasoning model itself with the capability to adaptively switch reasoning modes without introducing additional parameters. Other work like NoThinking~\citep{ma2025reasoning} instructs the model to skip reasoning by incorporating the special token (\texttt{</think>}) into the prompt. DualFormer~\citep{su2024dualformer} claims to achieve adaptive hybrid reasoning by dropping portions of reasoning steps during training. However, it only exhibits this capability on maze navigation and Sokoban tasks, failing on LLM reasoning tasks. Moreover, premature termination disrupts reasoning structure while random deletion removes critical content, degrading performance (\cf \Cref{tab:main_results}). In contrast, \MethodName\ preserves essential reasoning while removing redundant components, enabling adaptive reasoning without disrupting the reasoning chain. Additionally, ACPO~\citep{cheng2025incentivizing} employs reinforcement learning to optimize reasoning mode selection. However, this approach remains closed-source without available implementation. Moreover, it can only sequentially apply slow-thinking first then fast-thinking, rather than adaptively switching between the two modes.

\section{Conclusion}\label{sec:conclusion}
We propose \MethodName, a hybrid reasoning framework that integrates fast and slow thinking within a single large reasoning model. Specifically, \MethodName\ employs an auxiliary LLM-based judge to distinguish essential from redundant reasoning, constructing a hybrid fine-tuning dataset. This dataset is then used to fine-tune LRMs, equipping them with adaptive hybrid reasoning capabilities. Experiments on mathematical reasoning and question-answering tasks demonstrate that \MethodName\ significantly reduces reasoning tokens while maintaining competitive accuracy, validating the effectiveness of hybrid reasoning for efficient AI systems.

\section{Ethical Considerations}\label{sec:license-consider-ration}
All experiments in this work are conducted using open-source datasets and models, which does not involve any sensitive data and commercial use\footnote{\url{https://huggingface.co/datasets/tau/commonsense_qa} (MIT license)} \footnote{\url{https://github.com/allenai/OpenBookQA} (Apache 2.0 license)} \footnote{\url{https://huggingface.co/datasets/EleutherAI/asdiv} (CC-BY-4.0 license)} \footnote{\url{https://huggingface.co/datasets/openai/gsm8k} (MIT license)}. 10 senior NLP researchers were invited to voluntarily identify reasoning patterns, with full understanding of the research purpose. We refer readers to \Cref{sec:human-subjects} for more details

\section{Limitations}
While \MethodName demonstrates effective performance, it currently employs an auxiliary LLM-based judge to classify reasoning trajectories based on identified patterns. Future work could explore more end-to-end approaches where the model directly learns to distinguish essential from redundant reasoning through its internal representations, eliminating the need for external judging mechanisms.

\bibliography{reference}

\clearpage
\appendix

\section{Case Study on Redundant Reasoning}\label{appendix:redundant_reasoning}

In this section, we present three redundant reasoning cases as follows:

\subsection{Repeated Self-Validation}

\begin{figure}[h]
\centering
\footnotesize
\modelcomparison{
Joy can read 8 pages of a book in 20 minutes. How many hours will it take her to read 120 pages?
}{
<think>\ldots I think I've verified it enough ways. So, it's 5 hours. \ans{But just to recap} \ldots \ans{I double-checked it multiple ways,} which is helpful because sometimes when you do math \ldots Just to make sure, \ans{let me recap one more time}\ldots</think>
}{}{DeepSeek-R1-Distill-Qwen-7B (Generated Reasoning Tokens: 1416)}
\caption{Repeated Self-Validation on GSM8K.}
\label{fig:case-study-over-thinking-repeated-self-validation}
\end{figure}

 As shown in \Cref{fig:case-study-over-thinking-repeated-self-validation}, the LRM double-checks its own reasoning steps. It does this by going back over its previous conclusions, using phrases like ``I double-checked it multiple ways'' and ``let me recap one more time''. While this repeated checking helps ensure answers stay consistent, it also uses extra time and computational resources. Furthermore, once the model has already derived the correct answer, these extra checks are unnecessary.

\begin{figure}[h]
\centering
\footnotesize
\modelcomparison{%
Artemis is making tea for a party. She knows her mom drinks an 8-ounce cup of tea and uses one ounce of tea. She will use this same ratio for the party. The party has 12 people there and each of them wants a 6-ounce cup of tea. How many ounces of tea does she need?
}{
<think>\ldots Alternatively, \ans{maybe she is asking} how much tea she needs to add, not how much total tea is made? \ldots \ans{Wait, no, the question is}, how many ounces of tea does she need? \ldots \ans{But let me think if there's another interpretation.} Maybe the 8-ounce cup is the total tea, including the tea itself and water?\ldots</think>
}{}{DeepSeek-R1-Distill-Qwen-7B (Generated Reasoning Tokens: 1316)}
\caption{Defensive Assumptions on GSM8K.}
\label{fig:case-study-over-thinking-defensive-assumptions}
\end{figure}

\begin{figure}[h]
\centering
\footnotesize
\modelcomparison{%
Tim has 30 less apples than Martha, and Harry has half as many apples as Tim. If Martha has 68 apples, how many apples does Harry have?}{
<think>\ldots Is there \ans{another way} to approach this problem? \ldots \ans{Alternatively}, maybe I can visualize this with a diagram \ldots \ans{Another way} to think about it is by using equations \ldots</think>
}{}{ DeepSeek-R1-Distill-Qwen-7B (Generated Reasoning Tokens: 818)}
\caption{Multi-Solution Exploration on GSM8K.}
\label{fig:case-study-over-thinking-multi-solution}
\end{figure}

\subsection{Defensive Assumptions}
As shown in \Cref{fig:case-study-over-thinking-defensive-assumptions}, the LRM acts too cautiously by exploring multiple hypotheses. Specifically, it questions its initial understanding by expressing ``maybe she is asking'' or ``if there’s another interpretation''. Since the problem directly asks, ``How many ounces of tea does she need?''---a straightforward question requiring the total number, these additional hypotheses are unnecessary. It wastes additional computational resources, reducing the inference efficiency.

\subsection{Multi-Solution Exploration}
As illustrated in \Cref{fig:case-study-over-thinking-multi-solution}, the model continues to explore alternative solutions even after deriving the correct answer. This behavior is evidenced by recurring phrases such as `another way' `Alternatively' and `Another approach'. Since the model has derived the correct answer, exploring other solutions is redundant and wasting computational resources.

\clearpage

\section{Case Study on Essential Reasoning}\label{appendix:essential_reasoning}
In this section, we present three essential reasoning examples as follows:

\begin{figure}[h]
\centering
\footnotesize
\modelcomparison{%
Tony's dad is very strict about the washing machine and family members are only allowed to wash 50 total ounces of clothing at a time. Tony doesn't want to break the rules, so he weighs his clothes and finds that a pair of socks weighs 2 ounces, underwear weighs 4 ounces, a shirt weighs 5 ounces, shorts weigh 8 ounces, and pants weigh 10 ounces. Tony is washing a pair of pants, 2 shirts, a pair of shorts, and 3 pairs of socks. How many more pairs of underwear can he add to the wash and not break the rule?}{

<think>\ldots \ans{A pair of pants} weighs 10 ounces, \ans{each shirt} weighs 5 ounces, \ans{shorts} weigh 8 ounces, and \ans{each pair of socks} weighs 2 ounces \ldots Tony is washing \ans{one pair of pants}, \ans{two shirts}, \ans{one pair of shorts}, and \ans{three pairs of socks} \ldots </think>
}{}{DeepSeek-R1-Distill-Qwen-7B (Generated Reasoning Tokens: 214)}
\caption{Premise Omission Avoidance on GSM8K.}
\label{fig:case-study-essential-reasoning}
\end{figure}

\textbf{Premise Omission Avoidance}: The problem contains critical premises, and omitting any of these may lead to incorrect conclusions. As demonstrated in \Cref{fig:case-study-essential-reasoning}, the model systematically identifies and incorporates each premise during reasoning. Specifically, it explicitly catches: 1) the weight of each item, and 2) What Tony’s washing items includes.

\begin{figure}[htbp]
\centering
\footnotesize
\modelcomparison{%
One barnyard owl makes 5 hoot sounds per minute.  If 5 less than 20 hoots per minute are heard coming out of the barn, how many Barnyard owls would this noise be coming from?}{
<think>\ldots The problem states that this number is \ans{`5 less than 20.'} \ldots Next, I know that \ans{one barnyard owl makes 5 hoots per minute} \ldots</think>
}{}{DeepSeek-R1-Distill-Qwen-7B (Generated Reasoning Tokens: 154)}
\caption{Key-word Identification on GSM8K.}
\label{fig:case-study-essential-thinking-key-word}
\end{figure}

\textbf{Key-word Identification}: The model performs reasoning to identify and analyze critical keywords within the problem statement. In \Cref{fig:case-study-essential-thinking-key-word}, the model explicitly recognizes ``5 less than 20'' and ``one barnyard owl makes 5 hoots per minute'', which enables it to derive the correct solution. 

\textbf{Misunderstanding Prevention}: The model explicitly verifies task requirements to avoid misunderstanding. As shown in \Cref{fig:case-study-essential-thinking-misunderstanding}, when asked to calculate temperature decrease, the model recognizes this, rather than simply report the final temperature. It specifically highlights the phrases ``lose one-fourth of its initial temperature'' and ``find the decrease'', demonstrating awareness that the task requires computing the temperature change rather than deriving the resultant Addison mountain's temperature.

\begin{figure}[htbp]
\centering
\footnotesize
\modelcomparison{%
In one hour, Addison mountain's temperature will decrease to 3/4  of its temperature. If the current temperature of the mountain is 84 degrees, what will the temperature decrease by?}{
 <think>\ldots To determine the temperature decrease, I first recognize that the temperature decreases to three-fourths of its current value after one hour  \ldots This means the temperature will \ans{lose one-fourth of its initial temperature} \ldots Given that the current temperature is 84 degrees, I calculate one-fourth of this value to \ans{find the decrease}  \ldots </think>
}{}{DeepSeek-R1-Distill-Qwen-7B (Generated Reasoning Tokens: 94)}
\caption{Misunderstanding Prevention on GSM8K.}
\label{fig:case-study-essential-thinking-misunderstanding}
\end{figure}

\clearpage

\section{System Prompt}\label{appendix:system_prompt}

\begin{figure}[t]
        \centering
        \includegraphics[width=0.95\textwidth]{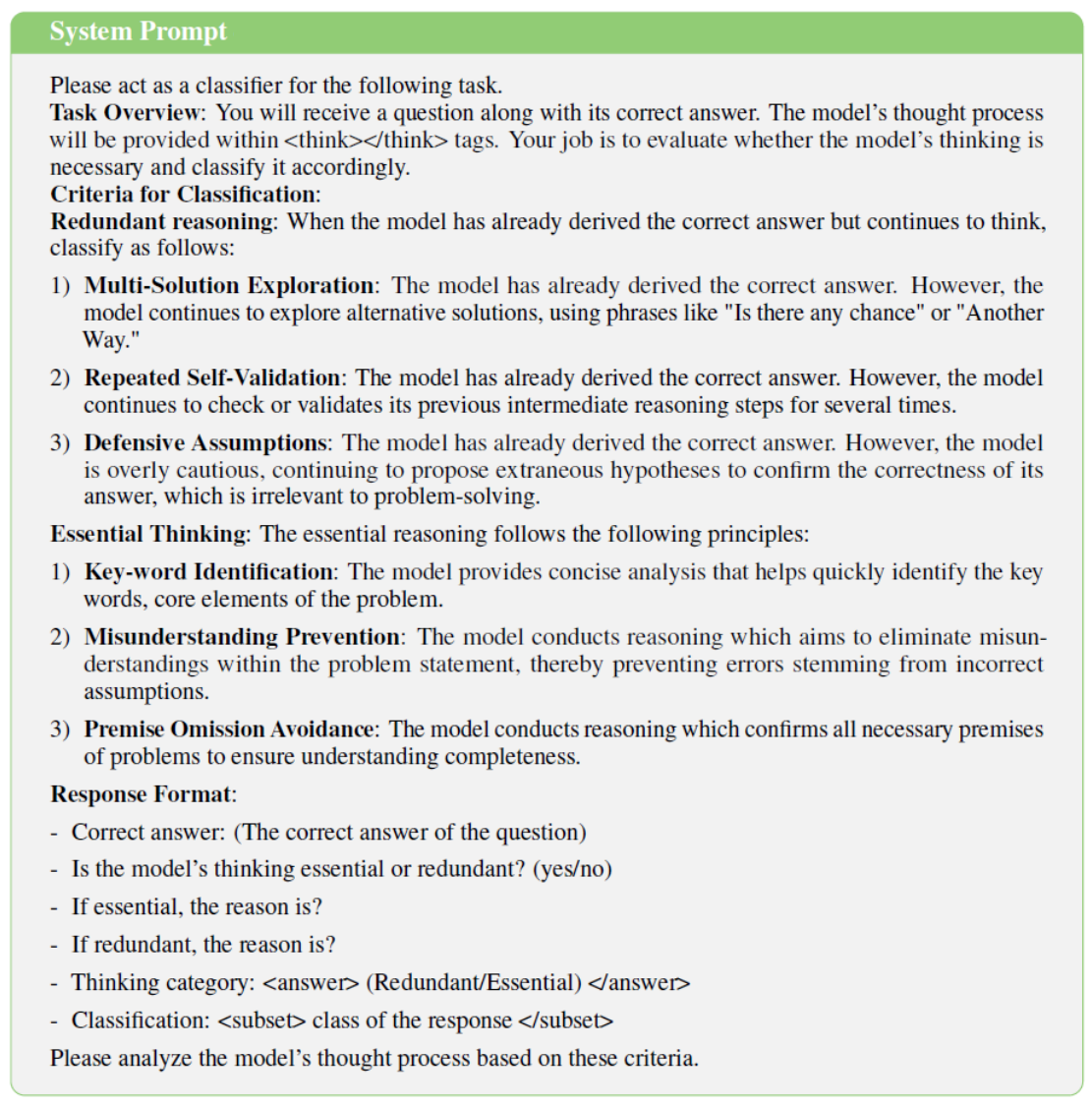}
            \captionsetup{justification=centering}
    \makebox[\textwidth][c]{%
        \parbox{0.9\textwidth}{\caption{System Prompt}\label{fig:system_prompt}}%
    }
\end{figure}

\MethodName\ employs an auxiliary LLM (GPT-4o) to distinguish essential from redundant reasoning. Based on the observed reasoning patterns, we construct prompts that guide the judge to classify reasoning trajectories, enabling systematic categorization of reasoning paradigms for hybrid dataset construction.

\clearpage

\section{Experimental Detials}
\subsection{Datasets}\label{appendix:experiment_datasets}

\noindent\textbf{Datasets.} We evaluated our approach using the following four datasets:
\begin{itemize}[leftmargin = *]
    \item OpenBookQA~\cite{OpenBookQA2018}: OpenBookQA provides the additional open-book knowledge for solving the multiple-choice questions. It consists of 4.96K training pairs with 500 validation and test examples. In this study, we intentionally exclude this open-book information during training and test, using only question-answer pairs. This design choice isolates the evaluation of \MethodName, as incorporating external knowledge risks conflating performance gains between retrieved knowledge and the method itself.
    
    \item CommonsenseQA~\cite{talmor-etal-2019-commonsenseqa}: CommonsenseQA consists of multiple-choice questions designed to evaluate models’ ability to apply diverse forms of commonsense knowledge in reasoning tasks. The dataset is split into 9.4K training, 1.22K validation and 1.14K test sets. As there is no answer key in the test set, the validation set is utilized to evaluate the performance of our proposed \MethodName. 

    \item ASDIV~\cite{miao2021diverse}: A dataset of elementary math word problems comprising 1,200 training and 301 test problems. The test set includes diverse problem types designed to assess basic arithmetic operations, such as aggregation, addition, and others.
    
    \item GSM8K~\cite{cobbe2021trainingverifierssolvemath}: GSM8K is a linguistically diverse dataset of 8.5K grade school math word problems—comprising 7.5K training and 1K test examples—that is designed to evaluate multi-step reasoning.
    
\end{itemize}

\subsection{SFT Dataset Construction}

We constructed two SFT datasets for training: a QA dataset combining OpenBookQA and CommonsenseQA and a MATH dataset merging ASDIV and GSM8K examples, with sample pruning ratios fully specified in \cref{tab:fast-thinking-ratio}. All original training instances were processed through DeepSeek-R1-Distill-Qwen ($\tau=0.9$, $\textit{top-p}=0.95$) to generate responses. The validation sets comprised 20\% held-out samples from each dataset except for OpenBookQA which has a predefined 500-sample validation set.

\subsection{Optimal Hyperparameters for Tuning }\label{appendix:optimal_hyperparameters}

\noindent\textbf{Hyperparameter Setting.} We use the open-source TRL library~\footnote{\url{https://github.com/huggingface/trl}} to train our models. The learning rate is searched in the range $\text{lr} \in \{ 3 \times 10^{-5}, 4 \times 10^{-5}, 5 \times 10^{-5}\}$, $\beta_1, \beta_2 \in \{10^{-3}, 10^{-4}\}$, \textbf{gas} (gradient accumulation steps) is searched in the range $\{1,4\}$, \textbf{batch} (batch size in each GPU device) is set as 2. The maximum padding length for SFT is set as 3000. For all experiments, we train for 4 epochs, saving a checkpoint after each epoch. Final reported performance corresponds to the model checkpoint (selected by epoch) and hyperparameters that achieve the highest validation accuracy. Training is conducted on 8 × 80GB NVIDIA A100 GPUs. 
The optimal hyperparameters for each method, across all datasets and model scales (7B, 14B), are listed in \cref{tab:hyperparameters-QA,tab:hyperparameters-MATH}, following the order of hyperparameters specified in \cref{tab:hyperparameters-searching}. We report the model results (\cf \Cref{tab:main_results}) with the best validation performance.

\noindent\textbf{Inference Setting.} For inference, we utilize the setting $\tau=0.9$, $\textit{top-p}=0.95$ for 7B/14B models. The LLM inference is conducted via the open-source vLLM library~\footnote{\url{https://github.com/vllm-project/vllm}}. 


\begin{table}[hb]
    \centering
    \caption{Hyperparameters to be searched for each method. \textbf{gas} denotes the \textit{gradient accumulation steps}; \textbf{batch} denotes the \textit{batch size in each GPU device}.}
    \label{tab:hyperparameters-searching}
    \begin{tabular}{l|l}
        \Xhline{1.2pt}
        \multicolumn{1}{c|}{\textbf{Method}} & \multicolumn{1}{c}{\textbf{Hyperparameters}} \bigstrut\\
        \Xhline{1pt}
        DualFormer   & \textbf{lr}, \textbf{gas}, \textbf{batch} \bigstrut\\
        \MethodName & \textbf{lr}, \textbf{gas}, \textbf{batch}, $\beta_1$, $\beta_2$ \bigstrut\\
        \Xhline{1.2pt}
    \end{tabular}
\end{table}

\begin{table*}[htbp]
    \centering
    \caption{Optimal hyperparameters and best-performing epoch checkpoint (selected by validation accuracy) on QA.}
    \begin{tabular}{c|l|cccccc}
    \Xhline{1.5pt}
    \textbf{Model Size} & \multicolumn{1}{c|}{\textbf{Method}} & \multicolumn{5}{c}{\textbf{Hyperparameters}} & \multicolumn{1}{c}{\textbf{Checkpoint Epoch}}  \bigstrut\\
    \Xhline{1pt}
        \multirow{2}[2]{*}{\textbf{ 7B}} 
            & DualFormer   & $3\times 10^{-5}$ & 4  & 2  &       &  & 2\bigstrut \\
            & \MethodName  & $3\times 10^{-5}$ & 4  & 2  &  $10^{-3}$     & $10^{-3}$ & 1\bigstrut \\
    \Xhline{0.8pt}
        \multirow{2}[2]{*}{\textbf{ 14B}} 
            & DualFormer   & $3\times 10^{-5}$ & 4  & 2  &       & & 2\bigstrut \\
            & \MethodName  & $3\times 10^{-5}$ & 4  & 2  &  $10^{-4}$     & $10^{-4}$ & 4\bigstrut \\
    \Xhline{1.5pt}
    \end{tabular}
    \label{tab:hyperparameters-QA}
\end{table*}

\begin{table*}[htbp]
    \centering
    \caption{Optimal hyperparameters and best-performing epoch checkpoint (selected by validation accuracy) on MATH.}
    \begin{tabular}{c|l|cccccc}
    \Xhline{1.5pt}
    \textbf{Model Size} & \multicolumn{1}{c|}{\textbf{Method}} & \multicolumn{5}{c}{\textbf{Hyperparameters}} & \multicolumn{1}{c}{\textbf{Checkpoint Epoch}}  \bigstrut\\
    \Xhline{1pt}
        \multirow{2}[2]{*}{\textbf{ 7B}} 
            & DualFormer   & $4\times 10^{-5}$ & 4  & 2  &       &  & 2\bigstrut \\
            & \MethodName  & $3\times 10^{-5}$ & 1  & 2  &    $10^{-3}$   &  $10^{-4}$& 2\bigstrut \\
    \Xhline{0.8pt}
        \multirow{2}[2]{*}{\textbf{ 14B}} 
            & DualFormer   & $3\times 10^{-5}$ & 4  & 2  &       &  & 4\bigstrut \\
            & \MethodName  & $3\times 10^{-5}$ & 1  & 2  &   $10^{-4}$    & $10^{-4}$ & 3\bigstrut \\
    \Xhline{1.5pt}
    \end{tabular}
    \label{tab:hyperparameters-MATH}
\end{table*}

\section{Potential Risks}
The training process was conducted on 8 $\times$ 80GB NVIDIA A100 GPUs, which involves environmental costs associated with energy consumption and carbon emissions. However, the proposed framework is designed to reduce the computational overhead of reasoning models during inference by decreasing token usage, which may reduce energy consumption across numerous inference calls.

\section{Details on Scientific Artifacts}
\noindent\textbf{Citation of artifact creators.} The creators of all scientific artifacts used in this work were properly cited. The experiments were conducted on four widely-used datasets: OpenBookQA~\citep{OpenBookQA2018}, CommonsenseQA~\cite{talmor-etal-2019-commonsenseqa}, ASDIV~\cite{miao2021diverse} , and GSM8K~\cite{cobbe2021trainingverifierssolvemath}. Each dataset was cited with reference to its original publication, and additional details regarding these datasets are provided in the appendix.

\noindent\textbf{License and terms of use.} The license and terms of use for the artifacts were considered in \Cref{sec:license-consider-ration}. All four datasets employed in this work—OpenBookQA, CommonsenseQA, ASDIV, and GSM8K—are publicly available and released under open-source licenses that permit research use.

\noindent\textbf{Artifact usage consistency.} The use of existing artifacts was consistent with their intended purposes. All four datasets—OpenBookQA, CommonsenseQA, ASDIV, and GSM8K—are standard benchmarks designed for research evaluation of question answering and mathematical reasoning capabilities, which aligns directly with their application in this work. These datasets were used solely for experimental evaluation purposes without modification. The proposed framework and any associated artifacts are intended exclusively for research purposes and academic evaluation, consistent with the original access conditions of the underlying datasets. 

\noindent\textbf{Privacy and content verification.} The datasets used in this work—OpenBookQA, CommonsenseQA, ASDIV, and GSM8K—are established benchmarks that have been curated and vetted by their original creators for research purposes. These datasets consist of question-answering and mathematical reasoning tasks that do not contain personal identifiable information such as names, addresses, phone numbers, or other sensitive data.

\noindent\textbf{Documentation of artifacts.}  Detailed descriptions of all four datasets used in this work—including their domains, sizes, linguistic characteristics, and intended evaluation purposes—are presented in \Cref{appendix:experiment_datasets}.

\section{Human Subjects}\label{sec:human-subjects}

\noindent\textbf{Instructions given to participants.} The analysis involved senior NLP researchers examining reasoning trajectories to identify patterns of essential reasoning. No formal experimental protocol with associated risks was required, as the task consisted solely of analytical review of model-generated text outputs without exposure to offensive content, personal identifying information, or other potentially harmful material. Researchers were asked to analyze reasoning trajectories, identify common patterns, and provide explanations for why these patterns characterize essential reasoning. Examples of the identified patterns are provided in \Cref{appendix:essential_reasoning,appendix:redundant_reasoning}.

\noindent\textbf{Recruitment and payment.} Participants were recruited from senior NLP researchers within the authors' academic network who volunteered to contribute their expertise to this analysis. No monetary compensation was provided, as participation was voluntary and based on academic collaboration.

\noindent\textbf{Data consent.} Prior to the involvement, participants were informed about the purpose of the study, specifically that their analytical expertise would be used to identify patterns in reasoning trajectories to guide the development of the proposed framework. Researchers were made aware that their analyses would contribute to characterizing essential versus redundant reasoning patterns in large reasoning models. All participants voluntarily agreed to contribute their expertise with full understanding of how their input would be utilized in the research.

\noindent\textbf{Ethics review.} The data collection protocol did not require ethics review board approval. The task involved technical analysis of model-generated text outputs by expert researchers and did not constitute human subjects research as defined by standard ethics review guidelines. No personal data, sensitive information, or experimental interventions involving human subjects were involved in this study.

\noindent\textbf{Characteristics of annotators.} This work did not involve human data annotation. The participating senior NLP researchers contributed analytical expertise to identify patterns in model-generated reasoning trajectories rather than performing data labeling tasks. No protected personal information or demographic data was collected, as the contribution was purely technical and analytical in nature.

\section{Acknowledgement on AI Assistant.} 

AI assistants were used in this research in the following capacities:

\begin{itemize}[leftmargin=*]
    \item \textbf{Research Methodology}: GPT-4o was employed as an LLM-based judge to classify reasoning trajectories as essential or redundant based on identified patterns (see \Cref{appendix:system_prompt} for details). This use is a core component of the proposed methodology and is fully documented in the paper.
    \item \textbf{Writing Assistance}: AI tools were used to assist with language refinement and editing of the manuscript, including grammar checking and stylistic improvements. All intellectual contributions, research design, experimental work, analysis, and core content generation were performed by the authors. The use of AI assistants was limited to editorial support and did not involve generation of scientific claims, results, or substantive content, in accordance with ACL publication ethics policies.
\end{itemize}

\end{document}